\documentclass[runningheads]{llncs}
\usepackage[T1]{fontenc}
\usepackage{graphicx}
\usepackage{amsmath}
\usepackage{multicol}
\usepackage{multirow}
\usepackage{hhline}
\usepackage{colortbl}
\usepackage{orcidlink}

\begin{document}
\title{From Persona to Person: Enhancing the Naturalness with Multiple Discourse Relations Graph Learning in Personalized Dialogue Generation}
\titlerunning{MUDI: Multiple Discourse Relations Graph Learning}
%
\author{
    Chih-Hao Hsu\inst{1} \and
    Ying-Jia Lin\inst{2}\orcidlink{0000-0003-4347-0232} \and
    Hung-Yu Kao\inst{2}\orcidlink{0000-0002-8890-8544}
}
\authorrunning{C.-H. Hsu et al.}
%

\institute{
    Department of Computer Science and Information Engineering,
    National Cheng Kung University, Taiwan \\
    \and
    Department of Computer Science,
    National Tsing Hua University, Taiwan \\
    \email{yingjialin@gapp.nthu.edu.tw}, \email{hykao@cs.nthu.edu.tw}
}

\maketitle              
\renewcommand{\thefootnote}{\fnsymbol{footnote}}
\footnotetext[1]{The first two authors contributed equally.}
\renewcommand{\thefootnote}{\arabic{footnote}}
\setcounter{footnote}{0}
\begin{abstract}
In dialogue generation, the naturalness of responses is crucial for effective human-machine interaction.
Personalized response generation poses even greater challenges, as the responses must remain coherent and consistent with the user's personal traits or persona descriptions.
We propose \textbf{MUDI} (\textbf{Mu}ltiple \textbf{Di}scourse Relations Graph Learning) for personalized dialogue generation.
We utilize a Large Language Model to assist in annotating discourse relations and to transform dialogue data into structured dialogue graphs.
Our graph encoder, the proposed DialogueGAT model, then captures implicit discourse relations within this structure, along with persona descriptions.
During the personalized response generation phase, novel coherence-aware attention strategies are implemented to enhance the decoder's consideration of discourse relations. Our experiments demonstrate significant improvements in the quality of personalized responses, thus resembling human-like dialogue exchanges.

\end{abstract}
\begin{figure}[ht]
    \centering
    \includegraphics[width=0.55\textwidth]{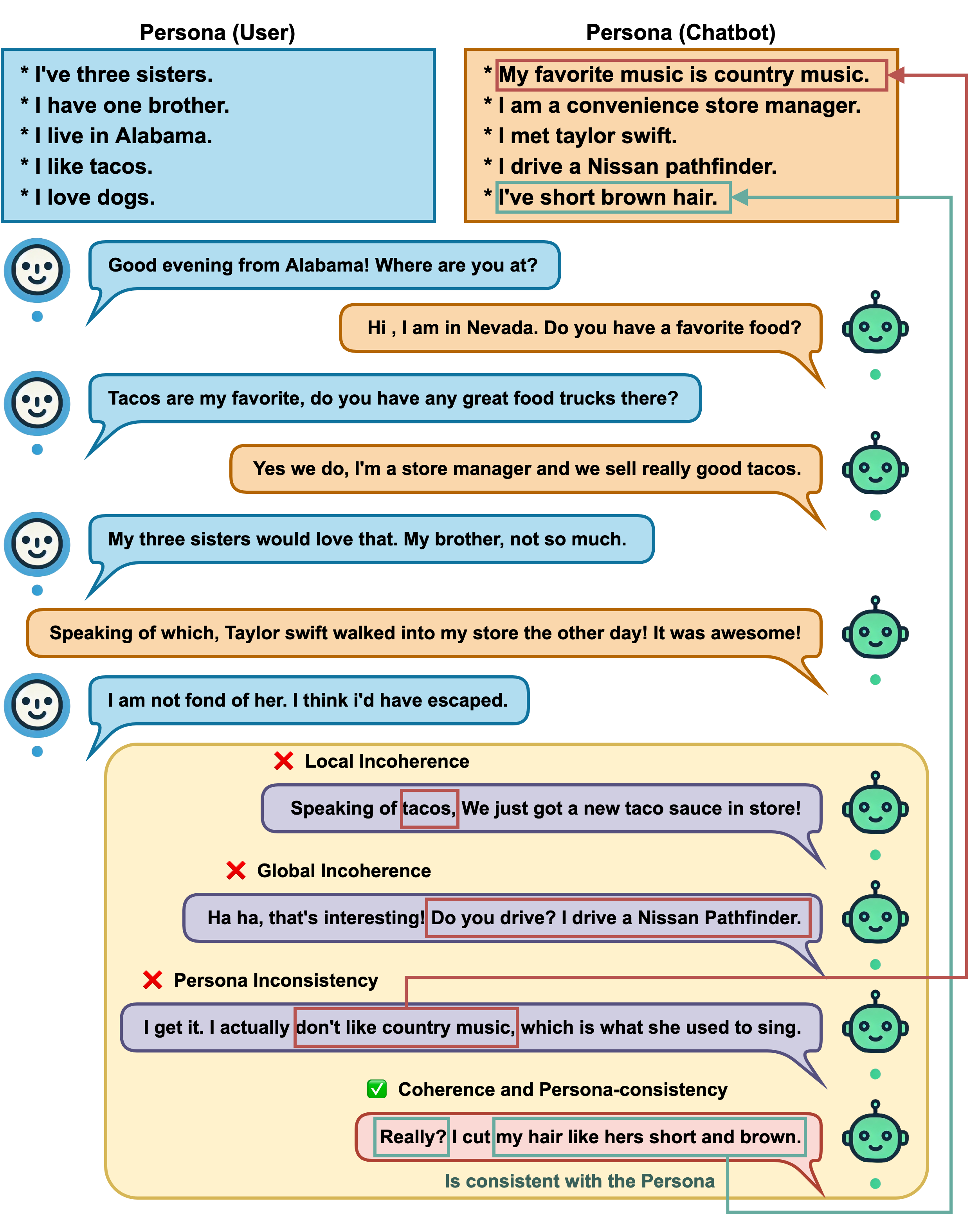}
    \caption{Example of Incoherence and Persona Inconsistency in a dialogue system.}
    \label{fig:intro_pdg}
\end{figure}
\section{Introduction}
Dialogue Generation is a foundational technology for dialogue systems, primarily focusing on the task of Next Utterance Generation, also known as Next Response Generation.
In multi-turn dialogue scenarios, the conversational agent's objective is to analyze the context of the multi-turn dialogue and the current query to produce a following appropriate response.
A significant drawback of traditional dialogue systems is their limited ability to personalize responses based on specific user characteristics or preferences. This limitation often results in generic and less engaging interactions that fail to meet individual user needs effectively \cite{jiang-de-rijke-2018-sequence}.
Previous works \cite{song-etal-2020-generating} defined this problem as the naturalness issue of the Dialogue System. One effective solution to enhance the naturalness of dialogue systems is to integrate personality into the agents, referred to as "persona". Typically, a persona comprises several sentences describing the interlocutor's facts or background. This information is crucial for building a trustworthy and confident dialogue system.
By endowing chatbot agents with human-like traits, the interactions become more realistic. Given these benefits, Personalized Dialogue Generation has emerged as a prominent research topic in recent years, focusing on improving user engagement and satisfaction within dialogue systems. This surge in interest is largely driven by the availability of large-scale personalized dialogue datasets, such as those from Zhang et al. \cite{zhang-etal-2018-personalizing} and Dinan et al. \cite{dinan-etal-2019-convai2}. These datasets have significantly advanced efforts to enhance persona consistency and context understanding in generated responses.
Innovative methods, including Liu et al. \cite{liu-etal-2020-impress} have concentrated on improving dialogue system consistency by modeling interlocutor understanding \cite{huang-etal-2023-paa,song-etal-2021-bob}.

Despite advances, challenges remain in enhancing engagement, coherence, and persona consistency.
The focus has been predominantly on the trade-offs between persona consistency and discourse coherence.
These challenges are primarily twofold. Firstly, many existing methods rely on sophisticated structures or external natural language inference (NLI) datasets to learn persona consistency.
This approach, while effective, can sometimes lead the model to overly prioritize persona information at the expense of neglecting the broader dialogue context. Secondly, many dialogue-generating models assume that that fluency alone can measure a dialogue's coherence and fail to adequately consider the importance of discourse relations.
Discourse coherence, which focuses on how utterances are interconnected and the overall organization of dialogue to effectively convey information, is essential for effective conversation. Discourse coherence can be divided into local and global coherence.
Local coherence refers to the logical connections between adjacent sentences, ensuring that they relate to each other and form a coherent sequence. Global coherence, on the other hand, extends beyond immediate sentence pairs to encompass higher-level relationships across the entire dialogue. This macro-linguistic capability allows conversational agents to maintain topic consistency and effectively convey meaning throughout an interaction.
Poor global coherence can significantly impair the user's understanding of the discourse as a cohesive whole.
As illustrated in Figure \ref{fig:intro_pdg}, the dialogue demonstrates various common issues encountered in personalized dialogue systems, including local and global incoherence as well as persona inconsistency. 

\begin{figure*}[ht]
    \centering
    \includegraphics[width=0.9\textwidth]{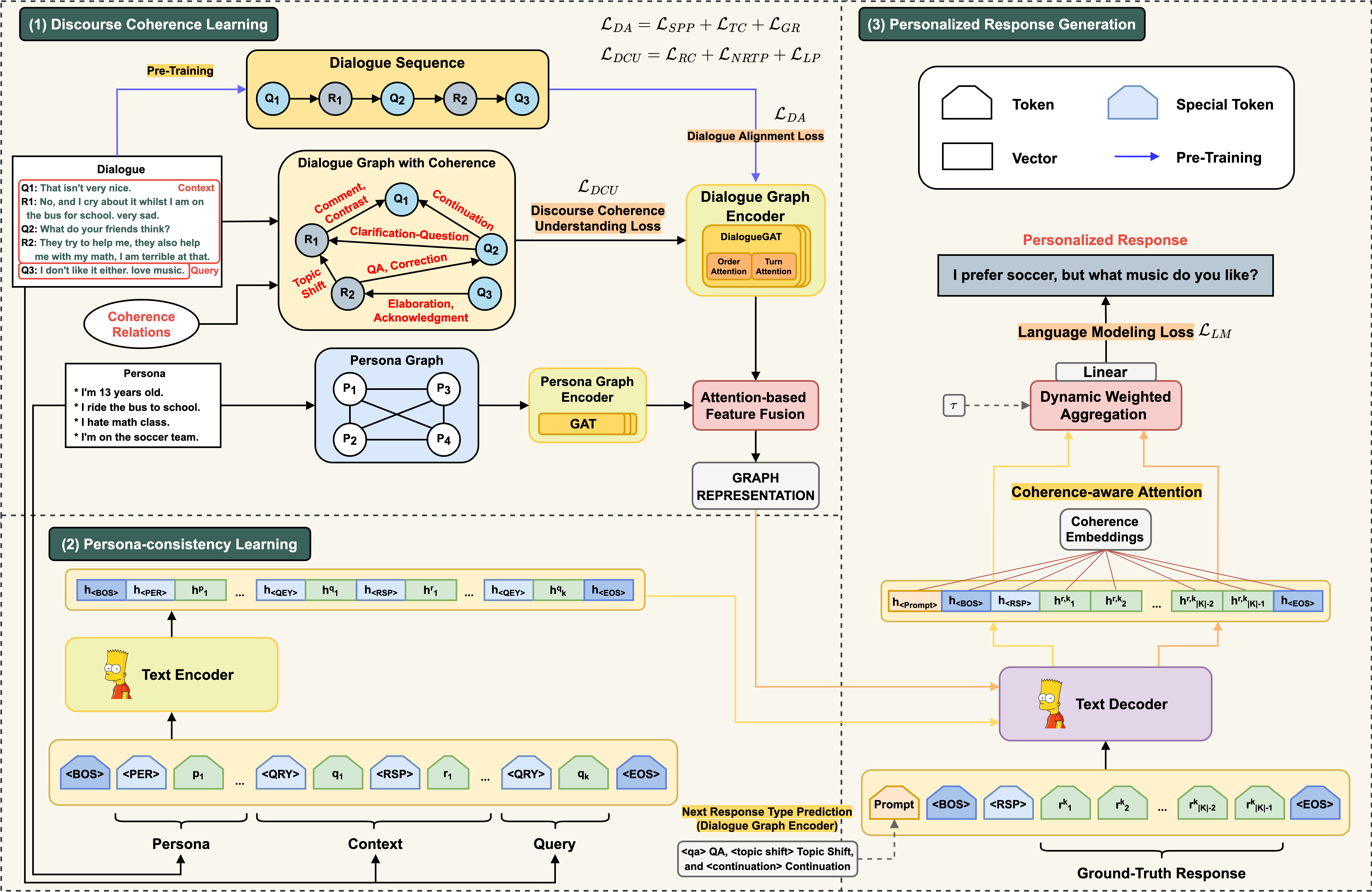}
    \caption{The overall architecture of our model - MUDI.}
    \label{fig:proposed_model_arch}
\end{figure*}

This study focuses on improving generation of responses that are coherent with the context and consistent with the persona, thus enhancing the naturalness of the personalized dialogue generation.
Our method is suitable for applications like customer service or healthcare assistants, where maintaining coherence and persona consistency is crucial for user trust.
We summarize our contributions:
\begin{itemize}
    \item \textbf{MUDI} is the first framework to jointly integrate Discourse Relations and Persona in Personalized Dialogue Generation, as shown in Fig. \ref{fig:proposed_model_arch}.
    
    \item Experiments on ConvAI2 \cite{dinan-etal-2019-convai2} show that \textbf{MUDI} achieves or surpasses baseline performance, generating more natural and personalized responses.
\end{itemize}

\section{Related Work}
The introduction of the PersonaChat dataset \cite{zhang-etal-2018-personalizing} marked a significant milestone of dialogue generation, which was expanded by the creation of the ConvAI2 dataset by Dinan et al. \cite{dinan-etal-2019-convai2} to serve as a key benchmark in persona-based dialogue tasks.
Jang et al. further enriched this domain by proposing a FoCus dataset that incorporates background knowledge alongside persona traits.
Before the advent of large personalized dialogue datasets \cite{zhang-etal-2018-personalizing}, researchers explored diversifying generated responses by incorporating speaker information into models by learning speaker embeddings.

For persona-related approaches, Zhang et al. employed LSTMs to fuse persona with contextual information \cite{zhang-etal-2018-personalizing}.
TransferTransfo fine-tuned GPT-2 with concatenated persona and dialogue inputs \cite{wolf-etal-2019-trans}, while BoB utilized three BERT models and the MNLI dataset to improve response relevance and consistency \cite{song-etal-2021-bob}.
P$^2$BOT introduced a unique architecture that enhances mutual persona perception \cite{liu-etal-2020-impress}.
Despite these advancements, maintaining persona consistency alongside coherence in responses remains a challenge.
A few studies like LMEDR, have attempted to address this by leveraging entailment and latent memory for discourse understanding \cite{chen-etal-2023-memorize}, yet the effectiveness for coherence is limited.
Therefore, while current methods align responses with personas, there is substantial scope for improving discourse coherence evaluation.

\section{Methodology}

\subsection{Task Formulation}
Given persona descriptions $P = \{p_{1}, p_{2}, \dots , p_{|P|}\}$ and a multi-turn dialogue context $C = \{q_{1}, r_{1}, q_{2}, r_{2}, ... , q_{|K|-1}, r_{|K|-1}, q_{|K|}\}$, where $q$ and $r$ denote user queries and chatbot responses, the task aims at generating a personalized response $r_{|K|}$.
The goal is to estimate the probability distribution $p(r | C, P)$ that incorporate persona information and dialogue history.
An ideal personalized response should be natural and consistent with the persona.
To ensure coherence, we incorporate the discourse relations in dialogue.
With specific response types $T = \{t_{1}, t_{2}, ... , t_{|T|}\}$ identified, we extend the objective to $p(r | C, P, T)$.

\subsection{Discourse Coherence Learning}

\subsubsection{Coherence Relations Annotation} \label{sec:coherence_reltaions_annotation}
To facilitate the model's understanding of coherence, we employ LLaMA-3-70B to assist in annotating coherence relations. According to STAC \cite{asher-etal-2016-discourse}, there are 16 discourse relations proposed. To these, we add a \textbf{topic-shift} to represent coherent topic transitions between conversations. Each pair of utterances can be annotated with up to three different relations.

\subsubsection{Dialogue Graph Modeling}
To capture discourse coherence in conversations for response generation, inspired by prior graph-based discourse modeling \cite{dong-etal-2021-discourse,feng-etal-2021-dialogue}, we use a graph encoder to learn the interactive relationships between discourses.
To account for sentence-level semantics, we use Sentence-BERT\footnote{https://huggingface.co/sentence-transformers/all-mpnet-base-v2} as an encoder to extract contextualized global semantics of utterances and personas, thereby initializing the node features.
Existing GNN models lack the design to fully capture dialogue structure and long-term interactions.
To overcome this, we enhance GATv2 \cite{brody-etal-2022-gatv2} with two key features: \textbf{Order information} and \textbf{Turn information}, integrated via attention mechanisms in our proposed \textbf{DialogueGAT}.

\subsubsection{Order-Attention}
To model the sequential nature of dialogues, we introduce auxiliary edges that connect each utterance to its $k{\text -}hop$ neighboring utterances based on their order. The existence of these edges is determined by the indicator function $I(i, j, d)$ (Eq. \ref{eq:add_order_auxiliary_edges}), where $d=k+1$ represents the maximum allowable order difference between connected nodes $i$ and $j$.

\begin{equation}\label{eq:add_order_auxiliary_edges}
I(i, j, d) = 
\begin{cases} 
1 & \text{if } \operatorname{order}(j) > \operatorname{order}(i) \\
  & \quad \text{and } |\operatorname{order}(i) - \operatorname{order}(j)| < d \\
0 & \text{otherwise}
\end{cases}
\end{equation}

The order-attention scores $s_{ij}$ between nodes $(i, j)$ are calculated based on the exponential decay of the order difference, as described in Eq. \ref{eq:dialoguegat_order_exp_decay} and \ref{eq:dialoguegat_order_act}.
\begin{equation}\label{eq:dialoguegat_order_exp_decay}
    s_{ij} = \exp(-\lambda \cdot |\operatorname{order}(i) - \operatorname{order}(j)|) \cdot I(i, j, d)
\end{equation}
\begin{equation}\label{eq:dialoguegat_order_act}
    e_{ij}^\mathrm{order} = (a^\mathrm{order})^\top \cdot \text{LeakyReLU}([W \cdot h_i  \parallel W \cdot h_j])) \cdot s_{ij}.
\end{equation}
Here, $a^\mathrm{order}$ and $W$ are learnable weights, and $\lambda$ controls the decay rate.

\subsubsection{Turn-Attention}
We also incorporate turn information by adding bidirectional auxiliary edges between utterance nodes within the same turn (Eq. \ref{eq:add_turn_auxiliary_edges}).

\begin{equation}\label{eq:add_turn_auxiliary_edges}
    t_{ij} = 
    \begin{cases} 
    1 & \text{if } \text{turn}(i) = \text{turn}(j) \\
    0 & \text{otherwise}
    \end{cases}
\end{equation}

Then, we calculate the turn-attention scores $t_{ij}$ between nodes of the same conversational turn similar to Eq. \ref{eq:dialoguegat_order_exp_decay} and \ref{eq:dialoguegat_order_act}:
\begin{equation}\label{eq:dialoguegat_turn_act}
    e_{ij}^\mathrm{turn} = ((a^\mathrm{turn})^\top \cdot \text{LeakyReLU}([W \cdot h_i  \parallel W \cdot h_j])) \cdot t_{ij}
\end{equation}
\noindent Since our model extends GATv2 \cite{brody-etal-2022-gatv2}, the attention function $\alpha$ is defined as:
\begin{equation}\label{eq:gatv2_edge}
    e_{ij} = (a^\top \cdot \text{LeakyReLU}([W \cdot h_i  \parallel W \cdot h_j]))
\end{equation}
\begin{equation}
    \alpha_{ij} = \mathrm{Softmax}(Concat(e_{ij}, e_{ij}^\mathrm{order}, e_{ij}^\mathrm{turn})).
\end{equation}

\subsubsection{Pre-training Phase}
Inspired by Wu et al. \cite{wu-etal-2023-gnn-pretrain}, we design three specific self-supervised pre-training tasks: \textbf{(1) Shortest Path Prediction}: Enables the model to predict the shortest paths between dialogue nodes, using Mean Squared Error loss ($\mathcal{L}_{\text{SPP}}$); \textbf{(2) Turn Classification}: Identifies whether successive utterances belong to the same speaker, optimized with Cross-Entropy loss ($\mathcal{L}_{\text{TC}}$); \textbf{(3) Graph Reconstruction}: Rebuilds dialogue sequences, also optimized with Cross-Entropy ($\mathcal{L}_{\text{GR}}$).
The pre-training loss  $\mathcal{L}_{\text{DA}}$ (\textbf{D}ialogue \textbf{A}lignment) is:
\begin{equation}
    \mathcal{L}_{\text{DA}} = \mathcal L_{\text{SPP}} + \mathcal L_{\text{TC}} + \mathcal L_{\text{GR}}
\end{equation}

\subsubsection{Fine-tuning Phase}
We use ConvAI2 \cite{dinan-etal-2019-convai2} annotated with coherence relations $R$ to refine the pre-trained graph encoder $\text{GNN}_{\theta}$.
Each node connects to its $k{\text -}hop$ neighbors, reflecting local conversational structures.
To address class imbalance in coherence relations,
we prune edges labeled exclusively with high-frequency categories to balance class distribution.
We update $\text{GNN}_{\theta}$ to $\text{GNN}_{\theta'}$ via:
\begin{equation}\label{eq:gnn_ft_enc}
    H_{\text{C}} = \text{GNN}_{\theta'}(X_{C}^{\text{ft}}, A_{C}^{\text{ft}}, R).
\end{equation}

In addition, we transform persona sentences from the dialogue into a complete graph for $\text{GNN}_{\psi}$ \cite{brody-etal-2022-gatv2},
which applies its attention mechanism across all connections to better capture the nuances and importance of each persona sentence.
\begin{equation}
H_{\text{P}} = \text{GNN}_{\psi}(X_{P}, A_{P}),
\end{equation}

\noindent A feature fusion is then performed for personalized node representations $H_{\text{D}}$:
\begin{equation}
\begin{aligned}
    &H_{\text{D}} = \text{MultiHeadAttention}(Q,K,V), \\
    &\text{where } 
    Q = H_{\text{C}} \cdot W^Q_i,
    K = H_{\text{P}} \cdot W^K_i, V = H_{\text{P}} \cdot W^V_i.
\end{aligned}
\end{equation}

\noindent Furthermore, coherence is learned through the three tasks:

\begin{itemize}
\item \textbf{Coherence Relations Classification}:
This multi-label classification task uses the graph encoder to predict the coherence relations exist of two nodes.
Cross-Entropy and Class-Balanced Loss are applied to \( \mathcal{L}_{\text{RC}} \).

\item \textbf{Next Response Type Prediction}:
This task predicts the next response type using node representations from $H_D$ to form a sequence $S$. Two methods are employed: (1) Direct Prediction, which uses the current utterance node, and (2) Sequential Prediction, which considers all prior utterances. The respective losses, $\mathcal{L}_{\text{NRTP}}^{\text{direct}}$ and $\mathcal{L}_{\text{NRTP}}^{\text{seq}}$, are calculated using Cross-Entropy.

\item \textbf{Link Prediction}:
Similar to Graph Reconstruction in pre-training, this task predicts whether an edge exists between two adjacent utterance nodes in the dialogue graph.
The model uses an inner product decoder for edge prediction, with the Cross-Entropy loss \( \mathcal{L}_{\text{LP}} \) applied to measure accuracy. 

\end{itemize}

In summary,
the total loss $\mathcal{L}_{\text{DCU}}$ (Discourse Coherence Understanding) of the dialogue graph encoder is the weighted sum of the losses:
\begin{equation}
    \mathcal{L}_{\text{DCU}} = \alpha \mathcal{L}_{\text{RC}} + \beta \mathcal{L}_{\text{NRTP}}^{\text{direct}} + \gamma \mathcal{L}_{\text{NRTP}}^{\text{seq}} + \delta \mathcal{L}_{\text{LP}},
\end{equation}
\noindent where \(\alpha\), \(\beta\), \(\gamma\), and \(\delta\) are the weights for the respective loss components.

\subsection{Persona-consistency Learning}
BART \cite{lewis-etal-2020-bart} is the backbone model for this stage and the subsequent personalized response generation.
Following \cite{chen-etal-2023-memorize}, the input to the text encoder is the concatenation of the persona descriptions $P$ and dialogue context $C$, which is structured as $[e_{\text{[BOS]}}, e_{\text{[PER]}}, e_{\text{p1}}, e_{\text{p2}}, ... , e_{\text{[QRY]}}, e_{\text{q1}}, e_{\text{[RSP]}}, e_{\text{r1}}, ... , e_{\text{[QRY]}}, e_{\text{q}|\text{K}|}, e_{\text{[EOS]}}]$.
The three special tokens, [PER], [QRY], and [RSP], indicate the beginning of persona, query, and response, respectively.

\subsection{Personalized Response Generation}
We adopt a prompt-based conditional dialogue generation approach, where the Prompt Tuning module guides the response generator.
Using predictions from the dialogue graph encoder about the next response type, the module generates descriptions incorporating response types, dialogue context, and persona.
The input sequence for the personalized response generator (Text Decoder) is structured as: 
$[e_{\text{[PROMPT]}}, e_{\text{[BOS]}}, e_{\text{[RSP]}}, e_{\text{1}}^{\text{k}}, e_{\text{2}}^{\text{k}}, ... , e_{|\text{K}|-1}^{\text{k}}, e_{|\text{K}|}^{\text{k}}, e_{\text{[EOS]}}]$.

To enhance coherence in next-token prediction, we apply cross-attention to dialogue representations from the dialogue graph encoder at each transformer block.
Each decoder layer attends to both standard encoder outputs and coherence-aware dialogue representations.
Furthermore, a coherence-aware attention mechanism with learnable embeddings captures semantic nuances of coherence relations by integrating special tokens into the prompt, guiding the generator to align with specific response types like Acknowledgment.
To balance coherence and persona consistency, inspired by \cite{huang-etal-2023-paa}, we build a Dynamic Weighted Aggregation mechanism,
which uses a threshold $\tau$ and a learnable mask to adjust the proportions of persona-aware and coherence-aware information.

For personalized response generation, we use language modeling to compute the probability distribution over the vocabulary for next-token generation.
Ultimately, the model learns to produce fluent, personalized, coherent, and persona-consistent responses with the dialogue history.
The loss function is defined as:
\begin{equation}
    \mathcal{L}_{\text{LM}} = - \sum_{t=1}^{T} \log P(y_t | y_{<t}, O_{\text{Generator}})
\end{equation}

\section{Experiments}
\subsection{Experimental Setup}
\subsubsection{Dataset}
We perform experiments on ConvAI2 \cite{dinan-etal-2019-convai2}, a chit-chat dataset based on PersonaChat \cite{zhang-etal-2018-personalizing}.
ConvAI2 comprises 17,878 and 1,000 multi-turn dialogues for the training and development sets, with a total of 131,438 and 7,801 utterances.
It features 1,155 unique training and 100 development persona descriptions.

\subsubsection{Baselines}
\textbf{(1) Personalized Dialogue Generation}: Strong models like BOB \cite{song-etal-2021-bob}, LMEDR \cite{chen-etal-2023-memorize}, and PAA \cite{huang-etal-2023-paa}.
\textbf{(2) General Dialogue Generation}: PLATO \cite{bao-etal-2020-plato} and DialoGPT \cite{zhang-etal-2019-dialogpt}.
For PLATO, following \cite{bao-etal-2020-plato}, we prepend the persona descriptions as part of the knowledge to the context during inference.
For DialoGPT, we adopt the post-processing approach as \cite{zhou-etal-2023-simoap}.
\textbf{(3) Large Language Model (LLM)}: GPT-4, tested with prompting for personalized response generation.

\subsubsection{Evaluation Metrics}
(1) \textbf{Text-similarity}:
We use BLEU \cite{papineni-etal-2002-bleu}, ROUGE \cite{lin-2004-rouge}, and BERTScore \cite{zhang-etal-2020-bert-score} to evaluate lexical overlap and semantic similarity.
(2) \textbf{Diversity}: Dist-$n$
\cite{li-etal-2016-diversity} and
Ent-$n$ \cite{zhang-etal-2018-generating} for unique n-grams and n-gram distribution,
and Unique Sentence Ratio (USR) \cite{li-etal-2020-generate} for sentence uniqueness.
(3) \textbf{Coherence}: We use QuantiDCE \cite{ye-etal-2021-towards-quantifiable} to assess dialogue coherence.
(4) \textbf{Personalization}: We use Consistency Score (C.Score) \cite{madotto-etal-2019-personalizing} based on predictions from an NLI model.
\begin{table*}[ht]
\centering
\def\arraystretch{1.0}%
 \resizebox{0.9\columnwidth}{!}{
\begin{tabular}{|c|c|c|c|c|c|c|}
\hline
\multicolumn{2}{|c|}{\multirow{2}{*}{\textbf{Model}}}  & \multicolumn{3}{|c|}{\textbf{Text Similarity}} & \multicolumn{1}{c|}{\textbf{Personalization}} & \multicolumn{1}{c|}{\textbf{Coherence}} \\
\cline{3-7}

\multicolumn{2}{|c|}{} & BLEU-1 $\uparrow$ & ROUGE-1 $\uparrow$ & BERTScore $\uparrow$ & C.Score $\uparrow$ & QuantiDCE $\uparrow$ \\
\hline

\rowcolor[RGB]{242,164,100}
\multicolumn{7}{|c|}{\textbf{Large Language Model (Prompting)}} \\
\hline

\multicolumn{2}{|c|}{GPT-4} &7.47 &13.52 &84.05 &2.86 &3.41 / 2.92 \\ 
\hline

\rowcolor{yellow}
\multicolumn{7}{|c|}{\textbf{General Dialogue Generation}} \\
\hline

\multicolumn{2}{|c|}{DialoGPT \cite{zhang-etal-2019-dialogpt}} &7.34 &9.46 &83.31 &4.53 &\textbf{3.23} / 2.79 \\ 
\hline

\multirow{2}{*}{PLATO \cite{bao-etal-2020-plato}} & w/ persona &4.35 &4.88 &82.77 &0.56 &1.68 / 1.57 \\ 
\cline{2-7}

\multirow{2}{*}{\textbf{}} & w/o persona  &6.82 &4.99 &81.44 &0.18 &1.87 / 1.77 \\ 
\hline

\rowcolor[RGB]{204,217,245}
\multicolumn{7}{|c|}{\textbf{Persona-based Dialogue Generation}} \\
\hline

\multicolumn{2}{|c|}{BoB \cite{song-etal-2021-bob}} &15.30 &13.21 &83.77 &0.51 &2.99 /2.76 \\ 
\hline

\multicolumn{2}{|c|}{LMEDR \cite{chen-etal-2023-memorize}}		&15.47	&13.28	&85.00 &7.83 &2.89 / 2.90 \\ 
\hline

\multicolumn{2}{|c|}{PAA \cite{huang-etal-2023-paa}}          &\underline{16.55}      &13.53   &84.42 &\textbf{15.19} &2.70 / \underline{2.93}   \\
\hline

\multirow{3}{*}{\textbf{\begin{tabular}{@{}c@{}}MUDI \\ (ours)\end{tabular}}} &SP\textsubscript{$\tau=0.2$}	&15.14	&14.87	&85.07 &\underline{11.87} &3.05 / 2.84	\\
\cline{2-7}

\multirow{3}{*}{} &Emb\textsubscript{$\tau=0.2$}   &\underline{16.55}  &\textbf{17.10}	&\underline{85.42} &9.70 &\textbf{3.23} / \textbf{2.94}	\\
\cline{2-7}

\multirow{3}{*}{} &SP+Emb\textsubscript{$\tau=0.2$}   &\textbf{18.19}	&\underline{16.59}	&\textbf{85.53} &9.75 &\underline{3.21} / 2.92 \\
\hline
\end{tabular}}
\caption{Our main results on ConvAI2 \cite{dinan-etal-2019-convai2}. \textbf{Bold}: best; \underline{Underline}: second best (GPT-4 is included as a reference and not considered in ranking).}
\label{table:main-results}
\end{table*}

\subsection{Main Results}
We report results on Text Similarity, Personalization, and Coherence evaluation in Table \ref{table:main-results}.
Specifically, MUDI's BLEU-1 and ROUGE-1 scores outperform existing methods by 1.64 and 3.57, respectively.
For Coherence evaluation, MUDI outperforms baselines in QuantiDCE, particularly in assessing the coherence between the query and response (left-side scores).
In addition, our model achieves comparable Coherence scores to DialoGPT, which focuses on general dialogue generation.
This indicates that our approach enables the model to generate responses with enhanced local coherence.
Furthermore, MUDI achieves excellent results in global coherence, which evaluates the coherence between the entire dialogue context and the response (right-side scores).

For Personalization evaluation, PAA significantly outperforms other methods in scores for Personalization.
Upon further examination, we found this is because PAA frequently generates sentences that are exact restatements of the persona description, often ignoring the relevance to the query.
As a result, its high scores in Personalization are be attributed to this tendency.
Excluding the special case of PAA, MUDI achieves excellent results in Personalization compared to other methods.
Combined with the previously discussed Coherence results, this demonstrates that our approach successfully balances discourse relations and persona, generating responses that effectively consider both aspects simultaneously.

Table \ref{table:text-diversity} highlights the Diversity evaluation, where MUDI generates completely unique responses across different queries and personas.
Additionally, MUDI achieves the highest Dist-1 scores among Persona-based dialogue generation methods, indicating its ability to produce varied and engaging responses.
\begin{table}[ht]
    \centering
    \def\arraystretch{0.8}%
    \begin{tabular}{|c|c|c|c|c|}
    \hline
    \multicolumn{2}{|c|}{\multirow{2}{*}{\textbf{Model}}}  &\multicolumn{3}{c|}{\textbf{Diversity}}\\
    \cline{3-5}
    
    \multicolumn{2}{|c|}{} &Ent-1 $\uparrow$ &Dist-1 $\uparrow$ &USR $\uparrow$ \\
    \hline
    
    \rowcolor[RGB]{204,217,245}
    \multicolumn{5}{|c|}{\textbf{Persona-based Dialogue Generation}} \\
    \hline
    
    \multicolumn{2}{|c|}{BoB \cite{song-etal-2021-bob}} &\underline{7.89} &41.75 &\underline{0.99} \\ 
    \hline
    
    \multicolumn{2}{|c|}{LMEDR \cite{chen-etal-2023-memorize}}  &7.14	&43.08	&0.94\\ 
    \hline
    
    \multicolumn{2}{|c|}{PAA \cite{huang-etal-2023-paa}}    &6.66	&40.27  &0.87 \\
    \hline
    
    \multirow{3}{*}{\textbf{\begin{tabular}{@{}c@{}}MUDI \\ (ours)\end{tabular}}} &SP\textsubscript{$\tau=0.2$}	&\textbf{8.13}		&46.76	&\textbf{1.00}\\
    \cline{2-5}
    
    \multirow{3}{*}{} &Emb\textsubscript{$\tau=0.2$}   &7.65	&\textbf{51.03}	&\textbf{1.00}\\
    \cline{2-5}
    
    \multirow{3}{*}{} &SP+Emb\textsubscript{$\tau=0.2$} 	&7.66	&\underline{47.68}  &\textbf{1.00}\\
    \hline
    \end{tabular}
    \caption{Diversity results on ConvAI2 \cite{dinan-etal-2019-convai2}. \textbf{Bold}: best; \underline{Underline}: second best.}
    \label{table:text-diversity}
\end{table}
\begin{figure}[ht]
    \centering
    \includegraphics[width=0.8\textwidth]{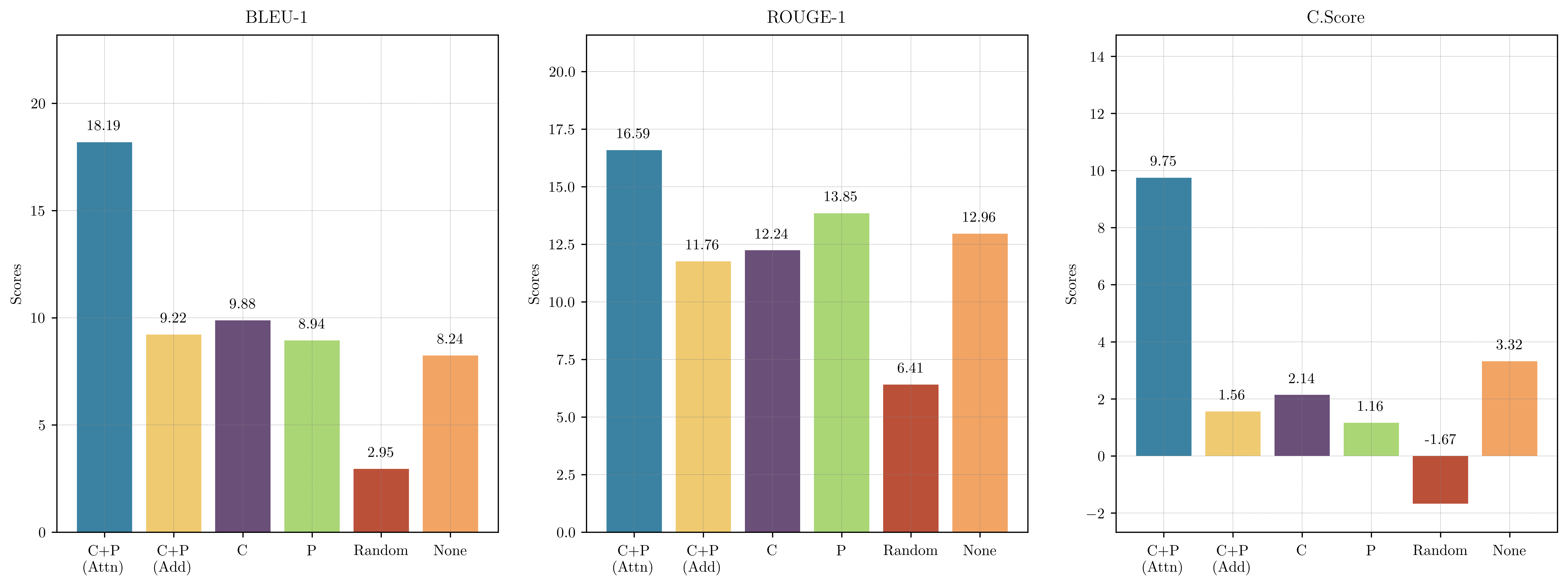}
    \caption{Performance analysis of dialogue graph encoder across different settings. Here, "C" represents Context, and "P" represents Persona.}
    \label{fig:effectiveness_of_dialogue_graph_encoder}
\end{figure}

\subsection{Analysis: The Effect of Dialogue Graph Encoder}

We evaluated the dialogue graph encoder under: (1) \textbf{Context+Persona (Attention)}, using attention-based feature fusion; (2) \textbf{Context+Persona (Add)}, with simple addition of representations; (3) \textbf{Context} only; (4) \textbf{Persona} only; (5) \textbf{Random} vector substitution; and (6) \textbf{None}, where the encoder is removed.
Figure \ref{fig:effectiveness_of_dialogue_graph_encoder} shows that the attention-based approach outperforms others in BLEU-1, ROUGE-1, and C.Score, demonstrating the importance of integrating context and persona.
Methods lacking integration or focusing on a single aspect reduce performance, underscoring the need for meaningful input for coherent responses.

\section{Conclusion}
\textbf{MUDI} is the first framework to jointly integrate Discourse Relations and Persona in Personalized Dialogue Generation.
The proposed DialogueGAT captures dialogue structure and contextual discourse relations, along with an attention-based feature fusion method to integrate context relations and persona information.
Next, a text encoder is employed to capture persona-aware representations, while a prompt-based conditional generation and coherence-aware attention further enhance the coherence in responses.
Finally, we use Dynamic Weighting Aggregation to balance the coherence-aware and persona-aware representations. 
Our experiments show that \textbf{MUDI} significantly improves personalized response quality, making them more coherent, informative, and human-like.

%
%
%
\bibliographystyle{splncs04}
\bibliography{aaai25}
\end{document}